\documentclass{article}
\usepackage{ijcai25}

\usepackage{times}
\usepackage{soul}
\usepackage{url}
\usepackage[hidelinks]{hyperref}
\usepackage[utf8]{inputenc}
\usepackage[small]{caption}
\usepackage{graphicx}
\usepackage{amsmath}
\usepackage{amsthm}
\usepackage{booktabs}
\usepackage{algorithm}
\usepackage[switch]{lineno}
\usepackage{amsfonts}
\usepackage{algpseudocode}
\usepackage{siunitx}

\title{CHIRPs: Change-Induced Regret Proxy Metrics for Lifelong Reinforcement Learning}

\author{
John Birkbeck$^1$
\and
Adam Sobey$^{1, 2}$\and
Federico Cerutti$^{3}$\and
Katherine Heseltine Hurley Flynn$^{4}$\And
Timothy J. Norman$^1$\\
\affiliations
$^1$University of Southampton\\
$^2$The Alan Turing Institute\\
$^3$University of Brescia\\
$^4$Thales UK RTSI\\
\emails
j.w.g.birkbeck@soton.ac.uk,
ajs502@soton.ac.uk,
federico.cerutti@unibs.it,
katherine.heseltineflynn@uk.thalesgroup.com,
t.j.norman@soton.ac.uk
}

\begin{document}

\maketitle

\begin{abstract}
    Reinforcement learning (RL) agents are costly to train and fragile to environmental changes. They often perform poorly when there are many changing tasks, prohibiting their widespread deployment in the real world. Many Lifelong RL agent designs have been proposed to mitigate issues such as catastrophic forgetting or demonstrate positive characteristics like forward transfer when change occurs. However, no prior work has established whether the impact on agent performance can be predicted from the change itself. Understanding this relationship will help agents proactively mitigate a change's impact for improved learning performance. We propose Change-Induced Regret Proxy (CHIRP) metrics to link change to agent performance drops and use two environments to demonstrate a CHIRP's utility in lifelong learning. A simple CHIRP-based agent achieved $48\%$ higher performance than the next best method in one benchmark and attained the best success rates in 8 of 10 tasks in a second benchmark which proved difficult for existing lifelong RL agents.
\end{abstract}

\section{The Value of Measuring Change}

Humans are experts at adaptation. When we detect changes in the world, we can predict their consequences and replan our behaviours accordingly. In contrast, Reinforcement Learning (RL) agents perform poorly when change occurs; their adaptation is a product of trial and error rather than anticipation. While RL agents have outperformed humans in controlled conditions, they require vast amounts of experience to do so, and rapidly lose their ability as conditions begin to vary ~\cite{towardscontinualrl,acontinuallearningsurvey}.

These weaknesses mean that RL agents are inapplicable to complex problems that they may face in the real world. For example, consider an aerial drone designed to monitor a hazardous area. The drone could suffer damage to sensors or actuators, be exposed to dangerous weather and adversaries, or be assigned an unforeseen mission. Pre-deployment training cannot prepare an agent for all the possible variations; scenarios like these demand human-like adaptability from vastly smaller sample sizes than are usually used during RL training.

Lifelong Reinforcement Learning (LRL) pursues human-like adaptation by designing agents to exhibit characteristics when the task or the environment changes: fast adaptation or fast remembering ~\cite{towardscontinualrl}, the avoidance of interference during learning ~\cite{samestatedifferenttask}, or incrementally learning upon prior knowledge ~\cite{progressiveneuralnetworks}. Benchmarks for these agents are designed as sequences of Markov Decision Processes (MDPs) with curated differences requiring one or more of these characteristics for high performance.

Understanding the kinds of MDP change that may occur is fundamental to agent design decisions and benchmarking design. Despite this, no work has constructed a systematic relationship between MDP change and agent performance. Understanding this relationship could help agents achieve these characteristics proactively rather than relying upon RL's trial-and-error nature. 

This paper introduces Change-Induced Regret Proxies (CHIRPs), a class of metrics that use changes in MDP components to `proxy measure' the expected performance impact for agents exposed to that change. To date, no previous work has investigated a proxy measurement of this kind. Correlation is established between one example CHIRP and agent performance in two environments. Finally, we introduce a simple agent design that uses CHIRPs to outperform all other agents tested, demonstrating the potential of these metrics for lifelong reinforcement learning.

\section{Related Work}
Metrics for reinforcement learning can be categorized as model- or performance-based. Model-based metrics are functions of the MDP components, such as the difference between two state spaces. Performance-based metrics are calculated from agents after exposure to an MDP; a simple example is an agent's episodic return, which is commonly used for agent evaluation. 

Performance-based metrics such as episodic return are the default choice ~\cite{acontinuallearningsurvey,towardscontinualrl} for evaluating reinforcement learners. Some performance metrics are purpose-built for lifelong learning ~\cite{cora,domainagnostic} to quantify forward transfer or catastrophic forgetting. Beyond agent evaluations, performance-based metrics are also an obvious choice for measuring an MDP change's impact; simply calculate an agent's performance before and after a change. However, this is prohibitively expensive for multiple changes due to the cost of training agents for every change that will be measured. 

Model-based metrics have been used to create state abstractions ~\cite{metricsfinitemdps,measuringdifferencefinitemdps}, invariant representations of state spaces ~\cite{invariantrepresentations}, to determine MDP similarity via world-models approximations ~\cite{rbmsmdpsimilarity}. These metrics often have restrictive limitations, such as requiring costly world-model training ~\cite{continualnoveltydetection}, assumptions of finite MDPs ~\cite{measuringdifferencefinitemdps} and the inability to estimate them from sampling. However, model-based metrics have been used more recently for context detection while avoiding these issues. Context detection techniques have used model-based metrics defined for continuous MDPs ~\cite{statisticalcontextdetection} or over world-models ~\cite{reactiveexploration,dealingwithnonstationarity} to determine when MDP changes occur. The purposes vary, from shaping exploration for more efficient learning ~\cite{reactiveexploration}, finding invariant representations ~\cite{continualstructurednonstationarity} or predicting an unseen task index ~\cite{statisticalcontextdetection}. However, model-based metrics are generally unsuitable for non-trivial problems.

To date, no work has attempted to construct a model-based proxy metric of a performance-based metric, which would allow a rapid assessment of the similarity between complex tasks.

\section{Preliminaries}
\label{sec:prelim}
The Reinforcement Learning (RL) problem is defined as a Markov Decision Process (MDP) $\mathcal{M} = \{\mathcal{S}, \mathcal{A}, \mathcal{R}, \mathcal{P}, \gamma\}$ with state space $\mathcal{S}$, action space $\mathcal{A}$, reward space $\mathcal{R}: \mathcal{S} \times \mathcal{A} \times \mathcal{S} \implies \mathbb{R}$, transition probability density function $p(s', r | s, a) \in [0, 1]$ and discount rate $\gamma \in [0, 1]$ ~\cite{suttonbarto}.

An RL agent is tasked with learning an action policy $\pi: \mathcal{S} \rightarrow \mathcal{A}$ which maximises the \textit{returns}, the discounted sum of future rewards $G_t = \sum_{k=0}^{\infty}{\gamma^{k} R_{t+k+1}}$ from the current timestep $t$.

Lifelong Reinforcement Learning allows an MDP's components\footnote[1]{We treat the discount rate $\gamma$ as a static component due to its common treatment as a tunable hyper-parameter.} to change in time: $\mathcal{M}_L(t) = \{\mathcal{S}(t), \mathcal{A}(t), \mathcal{R}(t), \mathcal{P}(t), \gamma\}$. These components may change discretely, continuously, or as a semi-continuous mixture. For any change, $\mathcal{M}_L(t)$ can always be represented as a sequence of static MDPs, even where a change occurs every timestep. We therefore use static MDPs $\mathcal{M}_i$ throughout this paper to denote each unique MDP that occurs as $\mathcal{M}_L(t)$ evolves in time.

In episodic RL, a policy $\pi$'s performance in MDP $\mathcal{M}_a$ can be judged by its expected returns over $s \in \mathcal{S}^0_a$, the set of possible initial states: $\mathbb{E}\big[G_t |\mathcal{S}^0_a, \pi\big]$. In Lifelong RL, the expected returns of a policy also depend upon the current components of $\mathcal{M}_L(t)$; for clarity below, we omit $\mathcal{S}^0_a$ as part of the definition of $\mathcal{M}_a$ and include $\mathcal{M}_a$ as an explicit condition of the expectation: $\mathbb{E}\big[G_t |\mathcal{M}_a, \pi\big]$.

\section{Scaled Optimal Policy Regret}\label{sec:sopr}

A natural choice for measuring the impact of an MDP change is regret, the drop in an agent's returns ~\cite{mahmud2016clustering,lifetimepolicyreuse}. Regret is commonly used in a single-MDP setting to compare the performance of different agents. In this setting, we compare an agent's performance before and after MDP change occurs. A change of MDP may affect the minimum or maximum returns bounds; for example, a wall may block a high-reward goal state from being visited, barring a direct comparison of regret. Therefore, the pre- and post-MDP change regrets must first be scaled by the returns bounds before comparison.

Motivated by this, we propose Scaled Optimal Policy Regret (SOPR) as the `target' metric for proxy measurement:

\begin{align}\label{eqn:sopr}
    & \text{SOPR}(\mathcal{M}_i, \mathcal{M}_j) = \notag \\
    & \sum_{\pi^+_i \in \Pi^+_i}\bigg[\frac{\mathbb{E}\big[G_t |\mathcal{M}_j, \pi^+_j \big] - \mathbb{E}\big[G_t |\mathcal{M}_j, \pi^+_i \big]}{\mathbb{E}\big[G_t |\mathcal{M}_j, \pi^+_j \big] - \mathbb{E}\big[G_t |\mathcal{M}_j, \pi^-_j \big]} \bigg], 
\end{align}
where $\pi^+_j$ is an optimal returns-\textit{maximising} policy of $\mathcal{M}_j$ and $\pi^-_j$ is an optimal returns-\textit{minimising} policy of $\mathcal{M}_j$. Note that $\text{SOPR}(\mathcal{M}_i, \mathcal{M}_j) \in [0, 1]$ for any pair of MDPs where SOPR is calculable. 

\subsubsection{When is SOPR calculable?}

By inspection of \eqref{eqn:sopr}, SOPR$(\mathcal{M}_i, \mathcal{M}_j)$ is defined if and only if the four expectations are defined. It is reasonable to assume three of the four terms are defined, as they are expectations of policies in their respective MDPs. This leaves only the expectation that `crosses' the MDPs, $\mathbb{E}\big[G_t |\mathcal{M}_j, \pi^+_i \big]$. 

Equation \eqref{eqn:sopr_calculable} is the standard definition ~\cite{suttonbarto} of expected returns for a single state $s_j \in \mathcal{S}_j$. For $\mathbb{E}\big[G_t |\mathcal{M}_j, \pi^+_i \big]$ to be defined, \eqref{eqn:sopr_calculable} must be defined for all initial states of $\mathcal{M}_j$. This implies that $\mathcal{P}_i(s', r | s, a)$ must be defined $\forall a \in \pi^+_i(a|s)$, and therefore that $\mathcal{A}_i \subseteq \mathcal{A}_j$. Additionally, $\pi^+_i(a|s)$ must be defined $\forall s' \in \mathcal{S}_j$, therefore $s' \in \mathcal{S}_j \implies s' \in \mathcal{S}_i$, i.e. $\mathcal{S}_i \subseteq \mathcal{S}_j$. 

Less formally, SOPR$(\mathcal{M}_i, \mathcal{M}_j)$ is calculable so long as $\mathcal{M}_i$'s optimal policies are executable in every state of $\mathcal{M}_j$.

\begin{align}\label{eqn:sopr_calculable}
    & \mathbb{E}\big[G_t |s_j, \pi^+_i \big] = \\ 
    & \sum_{a}\pi^+_i(a|s_j)\sum_{s_j', r}\mathcal{P}_i(s_j', r | s, a)\bigg[r + \gamma \mathbb{E}\big[G_{t+1} | s_j', \pi^+_i \big]\bigg]. \notag
\end{align}

\section{Change-Induced Regret Proxy (CHIRP) Metrics}

A desirable proxy for SOPR would have the following properties:
\begin{itemize}
    \item \textbf{Positively correlated} with SOPR: as the measured proxy distance between MDPs grows, the drop in performance should increase.
    \item \textbf{Monotonicity} with SOPR: a one-to-many mapping between a proxy and its target introduces undesirable uncertainty. Monotonicity avoids this.
    \item \textbf{Computational efficiency}: There is no utility in a proxy of a performance-based metric if it is also computationally expensive. 
    \item \textbf{Captures change across MDP components}: In lifelong RL we may encounter changes in any MDP component.
\end{itemize}

With these requirements in mind, we modify an existing metric to be used as an example CHIRP. In ~\cite{measuringdifferencefinitemdps}, a distance $d'(s_i, s_j)$ between states is defined in MDPs $\mathcal{M}_i$ and $\mathcal{M}_j$ as the maximum Wasserstein distance between every transition from that state. This state-to-state distance is aggregated over all state pairs to produce a distance between MDPs,
\begin{equation}\label{eqn:unsuitable_existing_metric}
    d(\mathcal{M}_i, \mathcal{M}_j) = \min_{\substack{k=1, \dots, |\mathcal{S}_1|, \\ t=1, \dots, |\mathcal{S}_2|}}\sum_{k=1}^{|\mathcal{S}_1|}\sum_{t=1}^{|\mathcal{S}_2|}\gamma_{kt}d'(s_k, s_t). 
\end{equation}

However, calculating this MDP distance is intractable for continuous state and action spaces and is difficult to estimate from samples; the minimum in \eqref{eqn:unsuitable_existing_metric} is taken over state spaces $\mathcal{S}_1$ and $\mathcal{S}_2$.

Instead, we use the Wasserstein distance between the transition distributions as a proxy for the MDP distance and propose its approximation via transition sampling. Generally, the 1-Wasserstein ($W_1$) distance between two probability distributions $\mathcal{P}_i(\mathbb{R}_d), \mathcal{P}_j(\mathbb{R}_d)$ is defined as,
\begin{equation}\label{eqn:wasserstein}
    W_1(\mathcal{P}_i, \mathcal{P}_j) = \inf_{\gamma \in \Gamma(\mathcal{P}_i, \mathcal{P}_j)} \int_{\mathbb{R}^d \times \mathbb{R}^d}||\mathbf{x} - \mathbf{y}||_2 \gamma(\delta\mathbf{x},\delta\mathbf{y}),
\end{equation}
where $\gamma \in \Gamma(\mathcal{P}_i, \mathcal{P}_j)$ is the set of all joint probability distributions with marginal distributions $\mathcal{P}_i$ and $\mathcal{P}_j$ ~\cite{slicedwasserstein}. Informally, the $W_1$ distance represents the minimal probability mass that must be moved from one distribution, $\mathcal{P}_i$, to produce another, $\mathcal{P}_j$.

Focusing upon MDPs, we define the $W_1$-MDP distance as the $W_1$ distance between transition probability density functions, 
\begin{align}
& W_1(\mathcal{M}_i, \mathcal{M}_j) = W_1(\mathcal{P}_i(s', r |s, a), \mathcal{P}_j(s', r |s, a))\notag  \\
& = \inf_{\gamma \in \Gamma(\mathcal{P}_i, \mathcal{P}_j)} \int_{\mathbb{R}^d \times \mathbb{R}^d}||\mathbf{s}^\ast_i - \mathbf{s}^\ast_j||_2 \gamma[\delta\mathbf{s}^\ast_i,\delta\mathbf{s}^\ast_j],\label{eqn:wasserstein_mdp}
\end{align}
with $\mathbf{s}^\ast_i = (\mathbf{s'}, r | \mathbf{s, a})_i$ distributed by $\mathcal{P}_i(\mathbf{s'}, r | \mathbf{s, a})$.

While $W_1(\mathcal{M}_i, \mathcal{M}_j)$ is still incalculable in continuous state and action spaces, its infimum is evaluated over the \textit{distributions} $\mathcal{P}_i$ and $\mathcal{P}_j$ rather than the state spaces as previously seen in \eqref{eqn:unsuitable_existing_metric}. This permits its estimation with empirical distributions $\hat{\mathcal{P}}_i$, $\hat{\mathcal{P}}_j$ calculated from transition samples. 

\section{Validating $W_1$-MDP as a CHIRP}

Before using the $W_1$-MDP distance as a CHIRP, we must establish whether it exhibits the desired characteristics listed above. It is unlikely that $W_1$-MDP is perfectly correlated with SOPR in every case, or that the relationship is the same in every environment, so understanding a CHIRP's limits is vital for its correct use.

One limitation is evident by inspection of SOPR and $W_1$-MDP; SOPR is `non-symmetric' for two MDPs as $\text{SOPR}(\mathcal{M}_i, \mathcal{M}_j) \neq \text{SOPR}(\mathcal{M}_j, \mathcal{M}_i)$ in general. However, for the Wasserstein distance, $W_1(\mathcal{P}_i, \mathcal{P}_j) = W_1(\mathcal{P}_j, \mathcal{P}_i)$ by definition. Although $W_1$-MDP is not a perfect substitute for agent performance, it may still be useful in practice. 

\subsection{Validation in SimpleGrid}

An ideal environment for investigating the link between CHIRP and SOPR would provide many MDP variants while being simple enough that both terms are easily calculated. This would require the total number of possible transitions to be relatively small, and the optimal policies to be easily determined without learning.

\begin{figure}[t]
  \centering
    \includegraphics[width=0.215\textwidth]{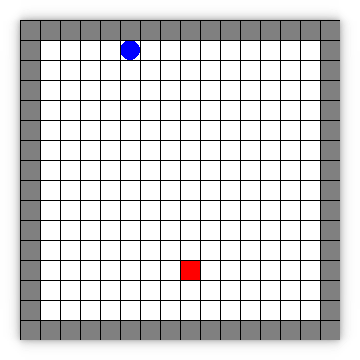}
    \caption{A SimpleGrid environment. The blue, circular agent must reach the red square goal. The grey edge squares are impassable walls.}\label{fig:simplegrid}
\end{figure}

To meet these requirements the `SimpleGrid' environment was constructed, inspired by MiniGrid ~\cite{MinigridMiniworld23}. SimpleGrid is a $20\times20$ grid where an agent can move in cardinal directions to reach a goal state. MDP variants are generated by varying the initial agent and goal positions. SimpleGrid has a 4-dimensional state space of the agent and goal $(x, y)$ positions and a discrete action space of length 4. The reward function is the negative scaled Manhattan distance to the goal:
\begin{equation}
    r(s, a, s') = -C|(x, y)_{\text{agent}} - (x, y)_{\text{goal}}|_1,
\end{equation}
with $C$ as a scaling constant determined by the initial agent and goal positions. Figure \ref{fig:simplegrid} illustrates the SimpleGrid environment. 

\begin{figure}[t]
  \centering
    \includegraphics[width=0.44\textwidth]{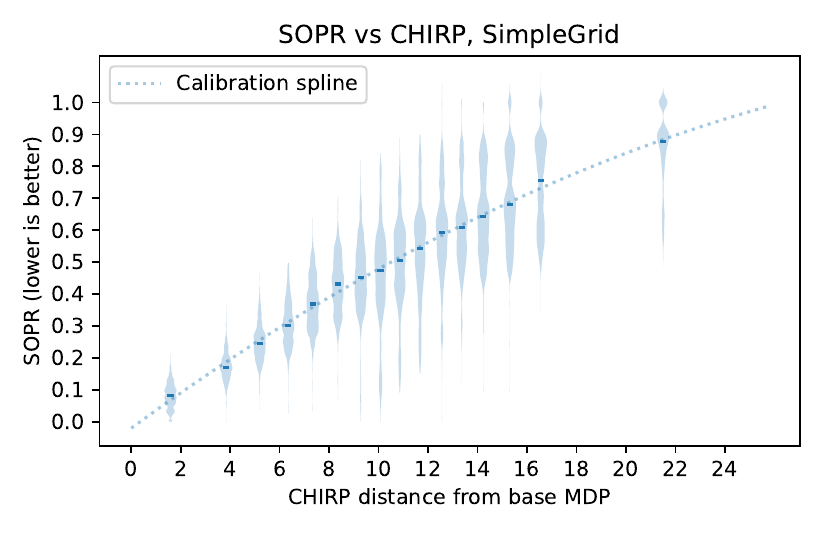}
    \caption{SOPR against our CHIRP for $10,500$ SimpleGrid MDP variants. The medians of each data bin are marked. Data was binned into 16 approximately equal volumes ($n\approx$ 650). The overlaid calibration curve is used further below.}
    \label{fig:010a_sg_chirp_sopr}
\end{figure}

Using SimpleGrid, the CHIRP and SOPR values for $\num{10500}$ MDP variations were calculated, as shown in Figure \ref{fig:010a_sg_chirp_sopr}. Visually, a positive monotonic relationship exists between SOPR and CHIRP. The Pearson correlation coefficient indicates that $W_1$-MDP is strongly positively correlated with SOPR $(\rho=0.875, p<0.001)$ and is strongly monotonic $(r_s=0.861, p<0.001)$ by Spearman's rank correlation.

Although this provides initial evidence for the proposed relationship, more complex environments like MetaWorld prohibit the analytical calculation of our CHIRP; we must first establish whether it can be estimated via sampling before moving beyond SimpleGrid.

\subsection{Approximating $W_1(\mathcal{M}_i, \mathcal{M}_j)$ with Sampling}\label{subsec:sampling}

Two schemes were considered for estimating $W_1(\mathcal{M}_i, \mathcal{M}_j)$, random and `reward-shaped' sampling. Both methods use sampled state-action pairs to generate transitions in each MDP; they differ in how the states and actions are sampled. As these methods estimate a proxy metric, the primary concern is the estimator's variance rather than bias, as calibration can remove bias.

In random sampling, states and actions are uniformly sampled from their respective spaces. In contrast, reward-shaped begins by uniformly sampling the MDPs' reward functions. The rewards are used as targets in the Monte Carlo Cross-Entropy ~\cite{mcce} method to generate state-action pairs whose transitions result in the sampled reward. Intuitively, reward-shaped sampling is intended to sample state-action pairs that are spread across low- and high-value states of the MDPs.

Both methods then execute the state-action pairs in $\mathcal{M}_i$ and $\mathcal{M}_j$, resulting in a transition sample from each MDP from the same state and action. Empirical distributions $\hat{\mathcal{P}}_i, \hat{\mathcal{P}}_j$ are constructed from these transition samples and are substituted with $\mathcal{P}_i, \mathcal{P}_j$ in Equation \eqref{eqn:wasserstein_mdp}.

In SimpleGrid, estimates of the previous $\num{10500}$ $W_1$-MDP distances were calculated under both sampling schemes from $15$ state-action pairs. In comparison, calculating the true value uses $\num{1296}$ state-action pairs. Both methods produced accurate estimates; mean errors were $-0.0025 \pm 0.0055$ for random sampling and $-0.0785 \pm 0.0081$ for reward-shaped sampling. Random sampling achieves a lower variance in SimpleGrid, marking it as the preferred choice.

\begin{figure}[t]
\centering
\includegraphics[width=0.45\textwidth]{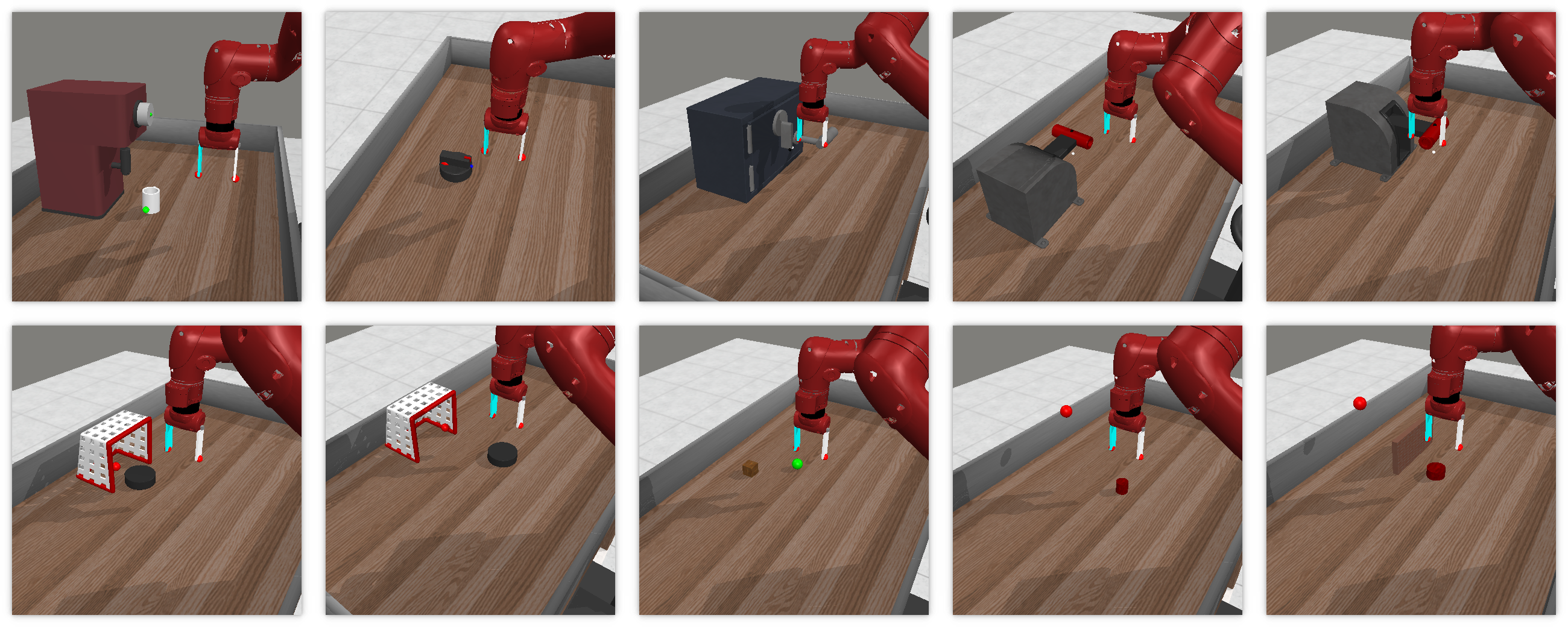}
\caption{The ten tasks selected for our experimentation in MetaWorld. Left to right, top to bottom: get coffee, turn dial, unlock door, press handle side, press handle, slide plate back, slide plate, push, reach, reach with wall.}\label{fig:metaworld}
\end{figure}

In Metaworld, 270 estimates were made using both methods across 10 MetaWorld tasks shown in Figure \ref{fig:metaworld}. Multiple samples sizes for states $n_s \in \{25, 50\}$ and transition repetitions $n_t \in \{1, 5, 10\}$ were used. Though the true CHIRP value is incalculable, the standard deviation of estimates can still be analysed: reward-shaped sampling achieved $0.3315$, $14\%$ smaller $(p<0.001)$ than random sampling's $0.3841$ with $95\%$ C.I.s $[0.3291, 0.3339]$ and $[0.3813, 0.3869]$ respectively. Reward-shaped sampling was therefore chosen for use in MetaWorld with $n_s = 50$ and $n_t = 1$. This sampling took $10.3$, $95\%$ C.I. [9.7, 10.8] seconds per MDP pair.

To generate transitions from sampled states and actions, the episodic reset functionality was extended to allow resetting to any state. Actions were then executed as normal to produce a transition. Resetting to specific states may be infeasible in the real world; in this case, $W_1$-MDP could be estimated from agent transitions. MDP sampling was used for this work to ensure that CHIRP estimates were not influenced by agent performance when analysing the CHIRP-SOPR relationship.

\subsection{Verification in MetaWorld}

To measure the CHIRP-SOPR relationship in MetaWorld, 20 soft actor-critic ~\cite{sac} agents per MetaWorld task were trained, giving 200 total agents. Each agent was exposed to increasing state and action errors within their respective tasks to simulate MDP change through sensor and actuator degradation. The SOPRs between the base and degraded MDPs were estimated using the highest and lowest returns observed across all agents.

\begin{figure}[t]
  \centering
    \includegraphics[width=0.44\textwidth]{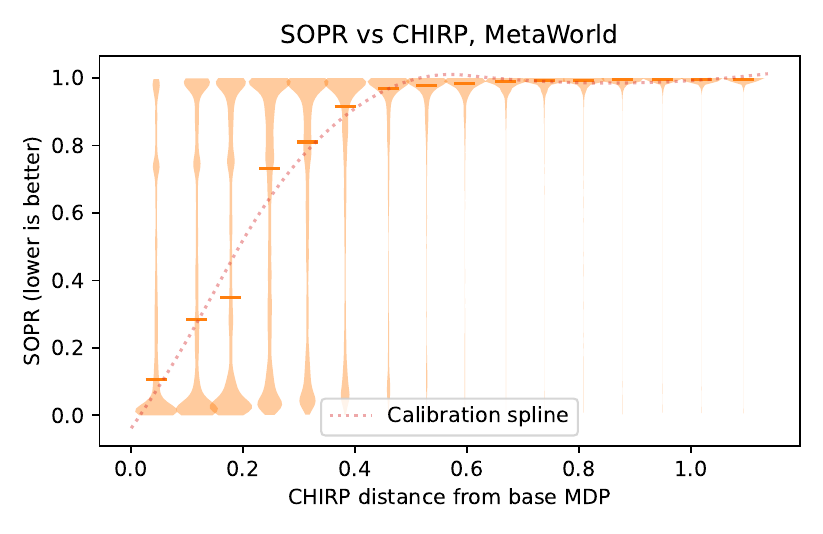}
    \caption{The CHIRP-SOPR relationship in MetaWorld for ten tasks. Each violin plot's median is marked. Data was binned to approximately equal volumes ($n\approx$ 5100). An example B-spline calibration curve is overlaid.}
    \label{fig:020a_mw_chirp_sopr}
\end{figure}

Figure \ref{fig:020a_mw_chirp_sopr} shows the aggregated CHIRP-SOPR relationship across the 200 agents and 10 tasks. A positive monotonic relationship exists with a moderate positive correlation $(\rho=0.60, p<0.001)$ and monotonicity $(r_s=0.70, p<0.001)$. The variance of SOPR values within each bin is wider than for SimpleGrid; this stems from the tendency of agents to either succeed or fail in MetaWorld, with fewer examples of `middling' performance after an MDP change. 

The relationship shown in Figure \ref{fig:020a_mw_chirp_sopr} is not linear, and total performance loss (a SOPR of 1) occurs at much lower CHIRP values than occurred in SimpleGrid. This illustrates how `raw' CHIRP measurements are incomparable across sufficiently different environments and motivates the example calibration further below.

\section{Lifelong Learning with a CHIRP}
\subsection{CHIRP Policy Reuse}
Analyzing a CHIRP's correlation with SOPR can establish whether a relationship exists but cannot inform us of the relationship's value. Is the correlation sufficient to be useful in lifelong reinforcement learning?

To test this, we modify \cite{lifetimepolicyreuse}'s Lifetime Policy Reuse (LPR) agent. In LPR, a multi-armed bandit learns a strategy to reuse $k$ policies across $n$ tasks with $k <= n$. The bandit and $k$ policies begin untrained and are learned with standard algorithms such as PPO ~\cite{ppo} and Q-learning ~\cite{suttonbarto}. 

Intuitively, LPR's bandit can be seen as finding groups of MDPs over which a single policy achieves good returns. Intuitively, this may also be achieved by finding clusters of MDPs with low CHIRP distances to one another. This could avoid the need to learn mappings and therefore the sub-optimal returns generated during a learning process.

\begin{figure}[t]
  \centering
    \includegraphics[width=0.4\textwidth]{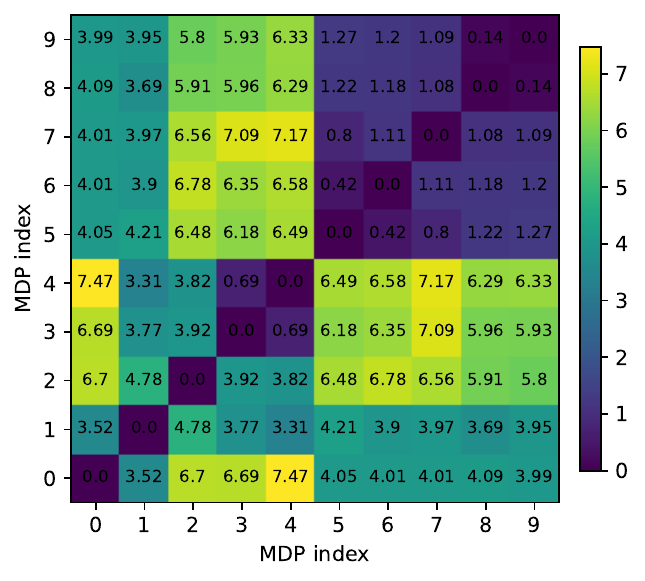}
    \caption{The median $W_1$ distances between 10 MetaWorld MDPs, shaded by distance value. The MDP index corresponds with Figure \ref{fig:metaworld}'s ordering; MDP 0 is `coffee button', and MDP 9 is `reach with wall'}
    \label{fig:mw_dist_matrix}
\end{figure}

We formalize this idea as CHIRP Policy Reuse (CPR). To pre-compute CPR's reuse strategy, the CHIRP values between the $\frac{n(n-1)}{2}$ unique MDP pairs are estimated, producing a distance matrix as in Figure \ref{fig:mw_dist_matrix}. $k$-medoids clustering ~\cite{kmedoids} is applied to this matrix to identify a clustering scheme that minimises within-cluster CHIRP values.

We compare CPR against LPR and two other LRL methods; \textit{Lifelong Reinforcement Learning with Modulating Masks} (Mask-LRL) ~\cite{modulatingmasks} and \textit{Lifelong Policy Gradient Learning of Factored Policies for Faster Training Without Forgetting} (LPG-FTW) ~\cite{lifelongpolicygradientlearning}, two approaches to lifelong learning whose methods are distinct from policy reuse. These methods were chosen based on their strong results in MetaWorld and Continual World ~\cite{continualworld}, a MetaWorld derivative.

Two experimental variants were used to test these agents, both of which used the 10 MetaWorld tasks used previously. For each method, 10 different seeded agents were trained. 

\subsection{Block learning of tasks}

In the block learning scenario, each task is seen once for $7 \times10^5$ timesteps each. Each agent's total success rate over the ten tasks was regularly evaluated to capture catastrophic forgetting, forward transfer and task capacity. This closely mimics the experimental design seen in Mask-LRL's original work ~\cite{modulatingmasks} but uses approximately $14\times$ fewer timesteps. This reduction impacts each method equally and increases the importance of sample efficiency.

In this scenario, Mask-LRL and LPG-FTW used hyperparameters defined in their original works and SAC used the hyperparameters used in MetaWorld SAC benchmarks ~\cite{metaworld}. The CPR and LPR methods both used the same SAC hyperparameters for each policy in their libraries.

\begin{figure}[t]
  \centering
    \includegraphics[width=0.44\textwidth]{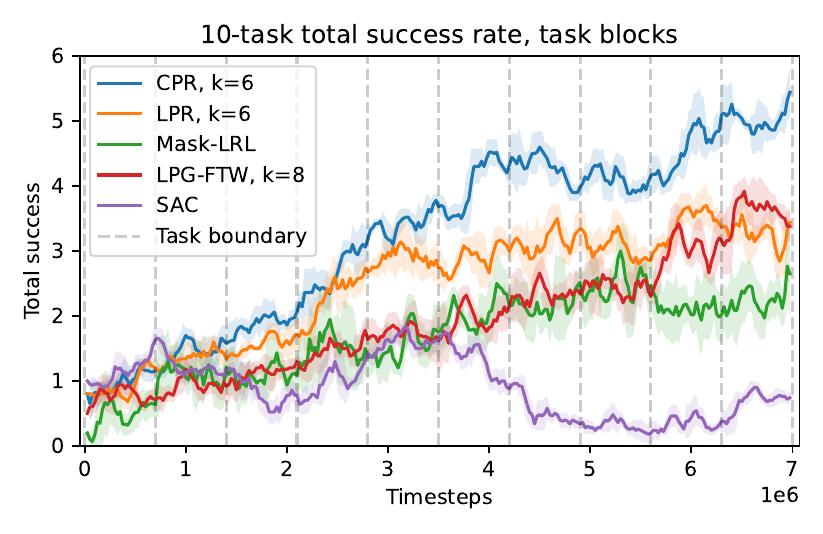}
    \caption{The rolling mean success rate across every task, with a window of 5 evaluations. Shaded areas indicate $95\%$ confidence intervals.}
    \label{fig:agent_comparison_14000}
\end{figure}

CHIRP Policy Reuse outperformed all other agents, achieving $30\%$ higher lifetime successes of $912$ versus LPR's $697$, with $95\%$ C.I.s of $[864, 960]$ and $[652, 742]$ respectively. In the final $\num{250000}$ timesteps, CPR achieved a mean total success of $52$, $48\%$ more than LPG-FTW's next best of $35$ with $95\%$ C.Is $[49.6, 54.4]$ and $[30.6, 39.4]$ respectively. Results for all agents are shown in Table \ref{tab:blocked_experiment} and Figure \ref{fig:agent_comparison_14000}.

\begin{table*}[t]
    \centering
    \begin{tabular}{l p{2.5cm} p{2.5cm} p{2.5cm} p{2.5cm} p{2.5cm} }
        \toprule
        Success criteria & \text{CPR} & \text{LPR} & \text{Mask-LRL} & \text{LPG-FTW} & \text{SAC} \\
        \midrule
        Lifetime         & \textbf{912}\hfill[864,\ 960] & 697\hfill[652,\ 742] & 596\hfill[562.0,\ 630]  & 544\hfill[496,\ 592]  & 252\hfill[228,\ 276] \\ 
        Last 250k steps  &\ \ \textbf{52}\hfill[50,\ 54] &\ \ 32\hfill[26,\ 38] &\ \ 33\hfill[29,\ 36]  &\ \ 35\hfill[31,\ 39]  &\ \ \ \  8\hfill[4,\ 11] \\ 
        \bottomrule
    \end{tabular}
    \caption{The total successes in the block learning scenario for each method. The sums of success rates were tested for Normality ~\protect\shortcite{normality} at the $5\%$ level before Normal C.I.s were calculated.}
    \label{tab:blocked_experiment}
\end{table*}

CPR's performance advantage over Mask-LRL and LPG-FTW provides strong evidence that our CHIRP metric has value for lifelong reinforcement learning. Although any CHIRP is imperfect, being only a proxy for performance, the simple usage in the CPR agent has outperformed multiple existing LRL methods with distinct designs.

It is particularly noteworthy that CPR outperformed Lifetime Policy Reuse; CPR is a simple replacement of LPR's bandit-learned mapping with a CHIRP-based clustering. This is surprising, as it is reasonable to expect a learned policy reuse strategy to outperform a fixed strategy based on a proxy for performance.

One cause for this result is the LPR agent's likelihood of becoming trapped in local optima. Table \ref{tab:clustering_vs_lpr_comp} compares one LPR agent's mapping against the CHIRP clustering result for $k=6$. The LPR agent has left one policy `spare' as at least one other policy outperforms it in every MDP. The spare is only selected during epsilon-greedy exploration; the infrequent training widens the performance gap and cements its exclusion.

\begin{table}[b]
    \centering
        \begin{tabular}{lp{0.05cm}p{0.05cm}p{0.05cm}p{0.05cm}p{0.05cm}p{0.05cm}p{0.05cm}p{0.05cm}p{0.05cm}p{0.05cm}}
        \toprule
        MDP index          & 0 & 1 & 2 & 3 & 4 & 5 & 6 & 7 & 8 & 9 \\
        \midrule
        $k$-medoid clustering & A & B & C & \textbf{F} & \textbf{F} & D & D & D & E & E \\
        LPR Strategy& A & B & C & \textbf{B} & \textbf{C} & D & D & D & E & E \\
        \bottomrule
        \end{tabular}
    \caption{An LPR agent's bandit-learned mapping neglects to assign policy F to any MDP. Numbers indicate MDPs, while letters represent the assigned policy.}
    \label{tab:clustering_vs_lpr_comp}
\end{table}

\subsection{Interleaved tasks}

Evaluating agents against every task at once is unrepresentative of their `true' performance, as they only experience one task at a time in the real world. Additionally, it is unlikely they experience one MDP for an extended period; it is more likely they can only attempt a task a handful of times before a change occurs. 

Motivated by this, the second experiment used random task changes that occurred with probability $10\%$ after each episode. Performance was assessed against the current task, rather than across all 10 at once. This harder setting placed additional focus on fast adaptation and remembering, as agents could not converge to each task's solution before a change occurred.

\begin{figure}[t]
  \centering
    \includegraphics[width=0.44\textwidth]{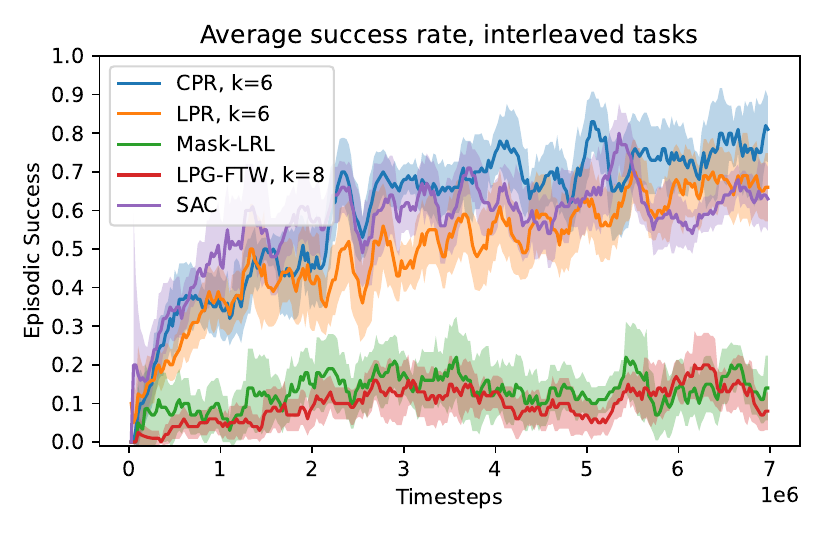}
    \caption{The windowed average success rate during training in the interleaved task benchmark. Wider 95\% C.I.s are caused by task change differences between each agent's runs, despite additional smoothing through a longer window (10 evaluation intervals)}
    \label{fig:agent_comparison_interleaved}
\end{figure}

CPR again outperformed all other methods significantly, achieving an average success rate of $79.7\%$ vs SAC's next best of $64.9\%$, a $22\%$ increase or $14.8$ percentage points higher. Table \ref{tab:agent_comparison_interleaved} records the performance figures for each method in the final $\num{250000}$ timesteps, and Figure \ref{fig:agent_comparison_interleaved} plots the running averages during training.

The per-task success rates of CPR in Table \ref{tab:agent_comparison_interleaved} are the highest of any agent in 8 out of 10 tasks and are above $97\%$ for 4 tasks. In particular, CPR drastically outperformed Mask-LRL and LPG-FTW, whose poor performance is likely due to the interleaved tasks and the increased importance of sample efficiency. In both methods, prior learning is reused via masking or policy composition for newly observed tasks, with an implicit assumption that previous task policies have converged before a new task is observed. This is invalidated in this setting. Additionally, this experimental design increases the importance of sample efficiency which is lower for these methods.

\begin{table*}[t]
    \centering
    \begin{tabular}{l p{2.5cm} p{2.5cm} p{2.5cm} p{2.5cm} p{2.5cm} }
        \toprule
        Success rate (\%) & \text{CPR} & \text{LPR} & \text{Mask-LRL} & \text{LPG-FTW} & \text{SAC} \\
        \midrule
        coffee button v2       & \textbf{99.2}\hfill[97.6,\ 99.7] & 85.3\hfill[81.2,\ 88.6] & 21.9\hfill[18.0,\ 26.5]  & 27.8\hfill[23.4,\ 32.6]  & 80.8\hfill[76.5,\ 84.6] \\ 
        dial turn v2           & \textbf{76.0}\hfill[71.8,\ 79.8] & 50.7\hfill[46.0,\ 55.4] & 13.6\hfill[10.7,\ 17.1]  &\ \ 5.1\hfill[3.4,\ 7.6]  & 56.7\hfill[52.0,\ 61.3] \\ 
        door unlock v2         & \textbf{90.4}\hfill[87.9,\ 92.4] & 68.2\hfill[64.6,\ 71.7] &\ \ 0.3\hfill[0.1,\ 1.1]  &\ \ 0.9\hfill[0.4,\ 2.0]  & 72.6\hfill[69.1,\ 75.8] \\ 
        handle press side v2   & \textbf{93.7}\hfill[91.0,\ 95.6] & 87.4\hfill[83.9,\ 90.2] & 25.1\hfill[21.2,\ 29.4]  & 32.6\hfill[28.3,\ 37.1]  & 72.8\hfill[68.4,\ 76.8] \\ 
        handle press v2        & \textbf{97.1}\hfill[95.2,\ 98.2] & 88.9\hfill[85.8,\ 91.3] & 39.8\hfill[35.7,\ 44.1]  &\ \ 1.0\hfill[0.4,\ 2.3]  & 92.2\hfill[89.5,\ 94.2] \\ 
        plate slide back v2    & 60.7\hfill[56.1,\ 65.1] & 31.1\hfill[27.0,\ 35.5] &\ \ 0.0\hfill[0.0,\ 0.8] &\ \ 0.2\hfill[0.0,\ 1.2]  & \textbf{77.9}\hfill[73.9,\ 81.5] \\ 
        plate slide v2         & \textbf{88.0}\hfill[84.8,\ 90.6] & 53.6\hfill[49.2,\ 58.0] &\ \ 8.1\hfill[6.0,\ 10.8] &\ \ 0.2\hfill[0.0,\ 1.2]  & 26.5\hfill[22.8,\ 30.6] \\ 
        push back v2           &\ \ 0.7\hfill[0.3,\ 1.9] &\ \ 3.1\hfill[1.9,\ 4.9] &\ \ 0.0\hfill[0.0,\ 0.7]  &\ \  0.0\hfill[0.0,\ 0.7] &\ \ \textbf{7.3}\hfill[5.4,\ 9.8] \\
        reach v2               & \textbf{97.2}\hfill[95.6,\ 98.2] & 89.4\hfill[86.7,\ 91.6] & 24.9\hfill[21.6,\ 28.5]  & 24.6\hfill[21.3,\ 28.2]  & 79.5\hfill[76.1,\ 82.6] \\
        reach wall v2          & \textbf{98.4}\hfill[97.0,\ 99.2] & 90.9\hfill[88.1,\ 93.1] & 31.5\hfill[27.7,\ 35.7]  & 20.4\hfill[17.2,\ 24.1]  & 85.6\hfill[82.3,\ 88.4] \\ 
        \midrule
        Overall:               & \textbf{79.7}\hfill[78.5,\ 80.8] & 64.7\hfill[63.3,\ 66.0] & 16.1\hfill[15.1,\ 17.1] & 10.6\hfill[9.7,\ 11.4] & 64.9\hfill [63.6,\ 66.2] \\
        \bottomrule
    \end{tabular}
    \caption{The interleaved scenario's per-task success rates over the final $\num{250000}$ timesteps with $95\%$ Wilson score confidence intervals ~\protect\shortcite{wilson}.}
    \label{tab:agent_comparison_interleaved}
\end{table*}

All agents struggled with `push back', though SAC surprisingly performed best. This task provides small rewards everywhere except when the task is solved, causing agents to learn much more slowly. This effect is so dramatic that agents trained in isolation for $\num{2000}$ episodes in `plate slide back' achieved a $14\%$ success rate in `push back', higher than the equivalent native agents' average of $4\%$.

This means that SAC is provided a surprising advantage by learning all tasks. A high-performance policy trained for other tasks has a higher chance of success in `push back' than an isolated policy trained poorly from sparse rewards. This leads to a positive feedback loop once high-return trajectories for `push back' enter the agent's replay buffer.

CPR benefits from this, but to a lesser degree. The `push back' task is grouped with `plate slide back' and `plate slide', but is only trained one-third as much as SAC's policy, whenever the current task is one of these three.

SAC's unexpectedly strong results also highlight a weakness in performing cross-task performance evaluations. Strong agent performance on the current task can be hidden if they perform poorly on hypothetical tasks. These evaluations are useful for measuring aspects of lifelong learning, but cannot be a replacement for actual performance during learning.

\section{Calibrating a CHIRP across environments}

Beyond the demonstrated use for learning, CHIRPs may have other uses in lifelong reinforcement learning. Results from existing lifelong benchmarks cannot be compared across environments, as their difficulties relative to one another are unknown. Without this, a consensus cannot form on which agent designs are best suited for lifelong learning \textit{generally}, not just for each benchmark individually. 

A calibrated CHIRP may help solve this problem. To demonstrate a simple calibration, b-splines were fit to the `raw' CHIRP-SOPR relationships, as seen in Figures \ref{fig:010a_sg_chirp_sopr} and \ref{fig:020a_mw_chirp_sopr}. 

\begin{figure}[t]
  \centering
    \includegraphics[width=0.44\textwidth]{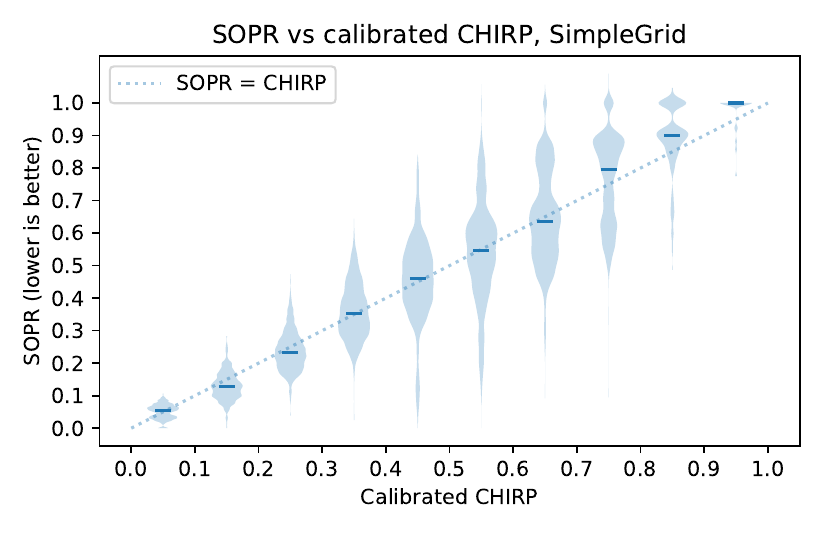}
    \caption{SimpleGrid's post-calibration CHIRP-SOPR relationship.}
    \label{fig:010b_sg_adj_chirp_sopr}
\end{figure}

\begin{figure}[t]
  \centering
    \includegraphics[width=0.44\textwidth]{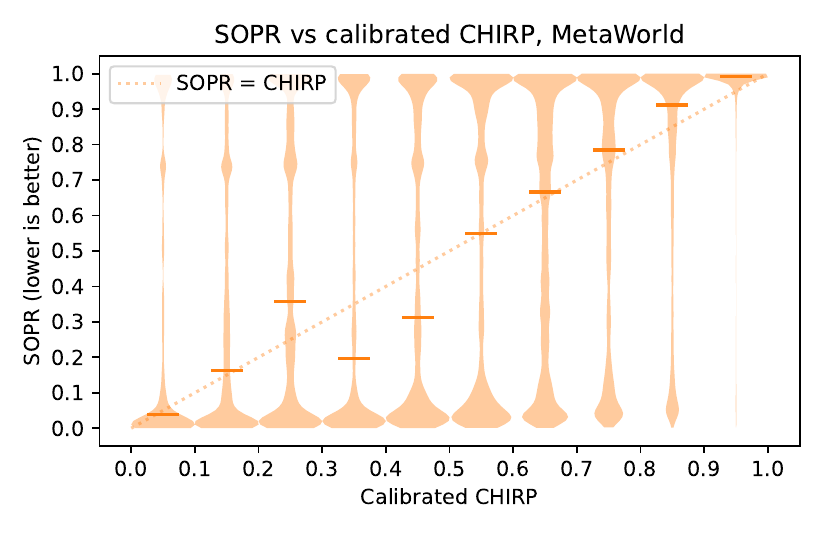}
    \caption{MetaWorld's post-calibration CHIRP-SOPR relationship.}
    \label{fig:020b_mw_adj_chirp_sopr}
\end{figure}

Figure \ref{fig:010b_sg_adj_chirp_sopr}'s results for SimpleGrid exemplify a desirable calibration; a calibrated value can be used as an estimate of SOPR directly. When combined with MetaWorld's calibrated results in Figure \ref{fig:020b_mw_adj_chirp_sopr}, the calibrated CHIRP values can be compared across environments to determine a predicted SOPR regardless of the base environment. Calibrating CHIRPs in this way permits novel comparisons of a change's difficulty that is agnostic to the benchmarking environments used.

\section{Conclusions}

Lifelong Reinforcement Learning studies the impacts of change on agents, but our understanding of this relationship is currently limited. Directly measuring agent performance across many changes is too costly, and model-based metrics are generally unsuitable for non-trivial problems.

This paper proposes CHIRP metrics and provides one example, the $W_1$-MDP distance. This was empirically shown to be positively correlated and monotonic with agent performance across two distinctly different environments. Unlike other model-based metrics, the $W_1$-MDP distance can capture change across every MDP component, is cheap to calculate, and does not assume that state or action spaces are discrete.

The value of the CHIRP concept for lifelong reinforcement learning was demonstrated with a novel agent design. This agent clusters MDPs by the example CHIRP to determine a policy reuse strategy. Despite this agent's simplicity, it outperformed all other agents tested in two experiments. The CHIRP Policy Reuse agent achieved 48$\%$ higher performance than the next best method in an existing lifelong RL benchmarking design. In the second experiment, the CHIRP based approach achieved the highest per-task success rates in 8 out of 10 tasks, and at least $97\%$ success rates in 4 tasks, well beyond other agents. Notably, this setting provided difficulties for two existing lifelong reinforcement learning agents.

\section*{Acknowledgements}
This work was supported by the UK Research and Innovation (UKRI) Centre for Doctoral Training in Machine Intelligence for Nano-electronic Devices and Systems [EP/S024298/1], the Lloyd’s Register Foundation and Thales UK RTSI. We also acknowledge the use of the IRIDIS High Performance Computing Facility and its support services at the University of Southampton in the completion of this work.

\appendix

\bibliographystyle{named}
\bibliography{ijcai25}

\end{document}